\pdfoutput=1

\documentclass[11pt]{article}

\usepackage[final]{acl}

\usepackage{times}
\usepackage{latexsym}

\usepackage[T1]{fontenc}

\usepackage[utf8]{inputenc}

\usepackage{microtype}

\usepackage{xcolor, soul}
\definecolor{ashgrey}{rgb}{0.80, 0.80, 0.80}
\sethlcolor{ashgrey}

\usepackage{graphicx}
\title{FrenchMedMCQA: A French Multiple-Choice Question Answering Dataset for Medical domain}

\author{
Yanis Labrak$^{1,4}$
\And Adrien Bazoge$^{2}$
\And Richard Dufour$^{2}$
\And Béatrice Daille$^{2}$
\AND Pierre-Antoine Gourraud$^{3}$
\And Emmanuel Morin$^{2}$
\And Mickael Rouvier$^{1}$
\AND \textnormal{\large LIA - Avignon University$^{1}$}\\\normalsize{\url{first.lastname@univ-avignon.fr}}
\And \textnormal{\large LS2N - Nantes University$^{2}$}\\\normalsize{\url{first.lastname@univ-nantes.fr}}
\AND \textnormal{\large CHU de Nantes - La clinique des données - Nantes University$^{3}$}
\And \textnormal{\large Zenidoc $^{4}$}\\
}

\begin{document}
\maketitle
\begin{abstract}
This paper introduces FrenchMedMCQA, the first publicly available Multiple-Choice Question Answering (MCQA) dataset in French for medical domain. It is composed of 3,105 questions taken from real exams of the French medical specialization diploma in pharmacy, mixing  single  and multiple answers.
Each instance of the dataset contains an identifier, a question, five possible answers and their manual correction(s). 
We also propose first baseline models to automatically process this MCQA task in order to report on the current performances and to highlight the difficulty of the task. A detailed analysis of the results showed that it is necessary to have representations adapted to the medical domain or to the MCQA task: in our case, English specialized models yielded better results than generic French ones, even though FrenchMedMCQA is in French. Corpus, models and tools are available online.
\end{abstract}

\section{Introduction}





Multiple-Choice Question Answering (MCQA) is a natural language processing (NLP) task that consists in correctly answering a set of questions by selecting one (or more) of the given $N$ candidates answers (also called {\it options}) while minimizing the number of errors. MCQA is one of the most difficult NLP tasks because it requires more advanced reading comprehension skills and external sources of knowledge to reach decent performance.


In MCQA, we can distinguish two types of answers: (1) single and (2) multiple ones. Most datasets focus on single answer questions, such as MCTest~\cite{richardson-etal-2013-mctest}, ARC-challenge~\cite{Clark2018ThinkYH}, OpenBookQA~\cite{OpenBookQA2018}, QASC~\cite{https://doi.org/10.48550/arxiv.1910.11473}, Social-IQA~\cite{https://doi.org/10.48550/arxiv.1904.09728}, or RACE~\cite{lai-etal-2017-race}.
To our knowledge, few studies have been done to construct medical MCQA dataset. We can cite the MedMCQA~\cite{pmlr-v174-pal22a} and HEAD-QA~\cite{vilares-gomez-rodriguez-2019-head} corpora which contain single answer questions in Spanish and English respectively. For the multiple answer questions, MLEC-QA~\cite{li-etal-2021-mlec} provides 136k questions in Chinese covering various biomedical sub-fields, such as clinic, public health and traditional Chinese medicine.


The French community has recently greatly increased its efforts to collect and distribute medical corpora. Even if no open language model is currently available, we can  cite the named entity recognition~\cite{neveol14quaero} and information extraction~\cite{grabar2018} tasks. However, they remain relatively classic, current approaches already reaching a high level of performance.

In this article, we introduce FrenchMedMCQA, the first publicly available MCQA corpus in French related to the medical field, and more particularly in the pharmacological domain. This dataset contains questions taken from real exams of the French  diploma in pharmacy. Among the difficulties related to the task, the questions asked may require a single answer for some and multiple ones for others. We also propose to evaluate state-of-the-art MCQA approaches, including an original evaluation of several word representations across languages.

Main contributions of the paper concern (1) the distribution of an original MCQA dataset in French related to the medical field, (2) a state-of-the-art approach on this task and a first analysis of the results, and (3) an open corpus, including tools and models, all available online.



\section{The FrenchMedMCQA Dataset}
\label{s:FrenchMedMCQA}

In this section, we detail the FrenchMedMCQA dataset and discuss data collection and distribution.


\subsection{Dataset collection}

The questions and their associated candidate answer(s) were collected from real French pharmacy exams on the remede\footnote{\url{http://www.remede.org/internat/pharmacie/qcm-internat.html}} website. This site was built around a community linked to the medical field (medicine, pharmacy, odontology...), offering multiple information (news, job offers, forums...) both for students and also professionals in these sectors of activity. Questions and answers were manually created by medical experts and used during  examinations. The dataset is composed of 2,025 questions with multiple answers and 1,080 with a single one, for a total of 3,105 questions. Each instance of the dataset contains an identifier, a question, five options (labeled from \texttt{A} to \texttt{E}) and correct answer(s). The average question length is 14.17 tokens and the average answer length is 6.44 tokens. The vocabulary size is of 13k words, of which 3.8k are estimated medical domain-specific words ({\it i.e.} related to the medical field). We find an average of 2.5 medical domain-specific words in each question (17\% of words in average of a question) and 2.0 in each answer (36\% of words in average of an answer). On average, a targeted medical domain-specific word is present in 2 questions and in 8 answers.


\subsection{Dataset distribution}

Table~\ref{table:ANSWERS} presents the proposed FrenchMedMCQA dataset distribution for the train, development (dev) and test sets  detailed per number of answers ({\it i.e.} number of correct responses per question). Globally, 70\% of the questions are kept for the train, 10\% for validation and last 20\% for testing.


\begin{table}[htb!]
\centering
 \resizebox{6.5cm}{!}{%
\begin{tabular}{c||c|c|c||c}
\hline
\textbf{\# Answers} & \textbf{Training} & \textbf{Validation} & \textbf{Test} & \textbf{Total} \\ \hline
\textbf{1}          & 595             & 164              & 321 & \textbf{1,080} \\
\textbf{2}          & 528             & 45               & 97 & \textbf{670} \\
\textbf{3}          & 718             & 71               & 141 & \textbf{930} \\
\textbf{4}          & 296             & 30               & 56 & \textbf{382} \\
\textbf{5}          & 34              & 2                & 7 & \textbf{43} \\ \hline\hline
\textbf{Total}      & \textbf{2171}   & \textbf{312}     & \textbf{622} & \textbf{3,105} \\ \hline
\end{tabular}%
 }
\caption{\label{table:ANSWERS}FrenchMedMCQA dataset distribution.}
\end{table}




\section{Methods}
\label{s:methods}

\begin{figure}[htb!]
\centering
\includegraphics[scale=0.23]{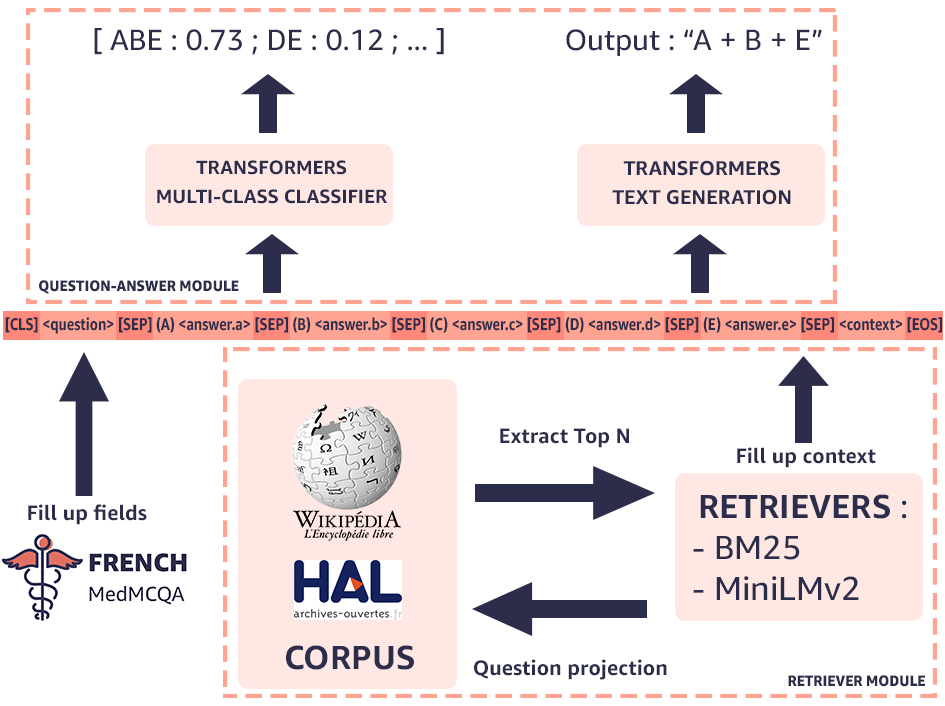}
\caption{Steps of the pipeline.}
\label{fig:Pipeline}
\end{figure}

The use alone of the question to automatically find the right answer(s) is not sufficient in the context of a MCQA task. State-of-the-art approaches then require external knowledge to improve system performances~\cite{izacard2020leveraging,https://doi.org/10.48550/arxiv.2005.00700}. In our case, we decide to build a two-step retriever-reader architecture comparable to UnifiedQA~\cite{https://doi.org/10.48550/arxiv.2005.00700}, where the retriever job is to extract knowledge from an external corpus and using it by the reader to predict the correct answers for each question. Figure~\ref{fig:Pipeline} presents the two-step general pipeline, first step being the retriever module, that extracts external context from the question (see Section~\ref{s:retriever}), and second step being the reader, called here question-answering module (see Section~\ref{s:qa}), that automatically selects answer(s) to the targeted question.

\subsection{Retriever module}
\label{s:retriever}

An external medical-related corpus fully composed of French has first been collected from two online sources: Wikipedia life science and HAL, the latter being an open archive run by the French National Centre for Scientific Research (CNRS) where authors can deposit scholarly documents from all academic fields. In our case, we focus on extracting papers and thesis from various specialization, such as Human health and pathology, Cancerology, Public health and epidemiology, Immunology, Pharmaceutical sciences, Psychiatric disorders and Drugs. This results in 1 million of passages ({\it i.e.} a portion of text that contains at least 100 characters) in HAL and 286k passages in Wikipedia.


This corpus is then used as a context extension for a question. We therefore used a retriever pipeline to automatically assign questions to the most likely passage in the external source. Two retrieval approaches are compared in this article: 

\begin{itemize}
\item BM25 Okapi~\cite{10.1145/2682862.2682863} for the implementation of the base BM25 algorithm~\cite{10.5555/106765.106783}.
\item SentenceTransformers framework~\cite{https://doi.org/10.48550/arxiv.1908.10084} is used to perform semantic search using state-of-the-art language representations taken from Huggingface’s Transformers library~\cite{https://doi.org/10.48550/arxiv.1910.03771}.
\end{itemize}


For both approaches, the goal is to embed each passage of the external corpus into a vector space using one of the two representations. On its side, the question is concatenated with the five options ({\it i.e.} answers associated to the question) to form a new query embedded in the same vector space. Embeddings from question and passages are finally compared to return the closest passages of a query (here, the cosine similarity is the distance metric). For the SentenceTransformers approach, we used a fast and non domain specific model called \texttt{MiniLMv2}~\cite{https://doi.org/10.48550/arxiv.2012.15828}. 
Note that the 1-best passage is only used in these experiments.

\subsection{Question-answer module}
\label{s:qa}

A goal of our experiments was to compare baseline approaches regarding two different paradigms. The first one is referred to a discriminative approach and consists in assigning one of $N$ classes to the input based on their projection in a multidimensional space. We also referred to it as a multi-class task. At the opposite, the second method is  a generative one which consists of generating a sequence of tokens, also called {\it free text}, based on a sequence of input tokens identical to the one used for the discriminative approach. The difference with the discriminative approach lies in the fact that we are not outputting a single class, like \texttt{ABE} for the question \texttt{6234176387997480960}, but a sequence of tokens following the rules of the natural language and referring to a combination of classes like \texttt{A + B + E} in the case of our studied generative model (see Section~\ref{s:generative}).

\subsubsection{Discriminative representations}

Four discriminative representations are studied in this paper. We firstly propose to use {\bf CamemBERT}~\cite{martin-etal-2020-camembert}, a generic French pre-trained language model based on RoBERTa~\cite{https://doi.org/10.48550/arxiv.1907.11692}. Since no language representation adapted to the medical domain are publicly available for French, we propose to evaluate the two pre-trained representations {\bf BioBERT}~\cite{Lee_2019} and {\bf PubMedBERT}~\cite{Gu_2022}, both trained on English medical data and reaching SOTA results on biomedical NLP tasks, including QA~\cite{pmlr-v174-pal22a}. Finally, we consider a multilingual generic pre-trained model, {\bf XLM-RoBERTa}~\cite{conneau-etal-2020-unsupervised} based on RoBERTa, to evaluate the gap in terms of performance with CamemBERT.

\subsubsection{Generative representation}
\label{s:generative}


Recently, generative models have demonstrated their interest on several NLP tasks, in particular for text generation and comprehension tasks. Among these approaches, {\bf BART}~\cite{https://doi.org/10.48550/arxiv.1910.13461} is a denoising autoencoder built with a sequence-to-sequence model. Due to its bidirectional encoder and left-to-right decoder, it can be considered as generalizing BERT and GPT~\cite{Radford2019LanguageMA}, respectively. BART training has two stages: (1) a noising function used to corrupt the input text, and (2) a sequence-to-sequence model learned to reconstruct the original input text. We then propose to evaluate this representation in this paper.

\section{Experimental protocol}
\label{s:experimental}

Each studied discriminative and generative model is fine-tuned on the MCQA task with FrenchMedMCQA training data using an input sequence composed of a question, its associated options ({\it i.e.} possible answers) and its additional context, all separated with a "[SEP]" token, e.g. \hl{[CLS] <question> [SEP] (A) <answer.a> [SEP] (B) <answer.b> [SEP] (C) <answer.c> [SEP] (D) <answer.d> [SEP] (E) <answer.e> [SEP] <context> [EOS]}. 


For each question, the context is the text passage with highest confidence rate and can either be obtained using the BM25 algorithm or semantic search as described in Section~\ref{s:retriever}.


Concerning the outputs of the systems, we have for the BART generative model a plain text containing the letter of the answers from $A$ to $E$ separated with plus signs in case of the questions with multiple answers, {\it e.g.} \hl{A + D + E}. For the other architectures ({\it i.e.} discriminative approaches), we simplify the multi-label problem into a multi-class one by classifying the inputs into one of the 31 existing combinations in the corpus. Here, a class may be a combination of multiple labels, {\it e.g.} if the correct answers are the $A$ and $B$ ones, then we consider the correct class being $AB$, which explains the number of 31 classes.

\begin{table*}[htb!]
\resizebox{\textwidth}{!}{%
\begin{tabular}{c||cc||cc|cc||cc|cc}
\hline
 & \multicolumn{2}{c||}{Without Context} & \multicolumn{2}{c|}{Wiki w/ BM25} & \multicolumn{2}{c||}{HAL w/ BM25} & \multicolumn{2}{c|}{Wiki w/ MiniLMv2} & \multicolumn{2}{c}{HAL w/ MiniLMv2} \\ \hline
Architecture & Hamming & EMR & Hamming & EMR & Hamming & EMR & Hamming & EMR & Hamming & EMR \\ \hline
BioBERT V1.1 & 36.19 & 15.43 & {\bf38.72} & 16.72 & 33.33 & 14.14 & 35.13 & 16.23 & 34.27 & 13.98 \\
PubMedBERT & 33.98 & 14.14 & 34.00 & 13.98 & 35.66 & 15.59 & 33.87 & 14.79 & 35.44 & 14.79 \\
CamemBERT-base & 36.24 & 16.55 & 34.19 & 14.46 & 34.78 & 15.43 & 34.66 & 14.79 & 34.61 & 14.95 \\
XLM-RoBERTa-base & 37.92 & 17.20 & 31.26 & 11.89 & 35.84 & 16.07 & 32.47 & 14.63 & 33.00 & 14.95 \\\hline
BART-base & 31.93 & 15.91 & 34.98 & {\bf18.64} & 33.80 & 17.68 & 29.65 & 12.86 & 34.65 & 18.32 \\ \hline
\end{tabular}%
}
\caption{\label{table:ResultsTableHammingScore}Performance (in \%) on the test set using the Hamming score and EMR metrics.}
\end{table*}

\subsection{Evaluation metrics}

The majority of tasks concentrate either on multi-class or binary classification since they have a single class at a time. However, occasionally, we will have a task where each observation has many labels. In this case, we would have different metrics to evaluate the system itself because multi-label prediction has an additional notion of being partially correct.
Here, we focused on two metrics called the Hamming score (commonly also multi-label accuracy) and Exact Match Ratio (EMR).

\subsubsection{Hamming score}

The accuracy for each instance is defined as the proportion of the predicted correct labels to the total number (predicted and actual) of labels for that instance. Overall accuracy is the average across all instances. It is less ambiguously referred to as the Hamming score rather than Multi-label Accuracy.

\subsubsection{Exact Match Ratio (EMR)}

The Exact Match Ratio (EMR) is the percentage of predictions matching exactly the ground truth answers. To be computed, we sum the number of fully correct questions divided by the total number of questions available in the set. A question is considered {\it fully correct} when the predictions are exactly equal to the ground truth answers for the question ({\it e.g.} all multiple answers should be correct to count as a correct question).


\section{Results}
\label{s:results}

Table~\ref{table:ResultsTableHammingScore} compiled the performance (in terms of Hamming score and EMR) of all the studied architectures and retrievers pipelines. For sake of comparison, the column {\it Without Context} has been added, considering that no retriever is used ({\it i.e.} no external passage is present in the QA system).

As we can see, the best performing model is different according to the used metric. \textbf{BioBERT V1.1} reaches best performance using the Hamming score and \textbf{BART-base} in the case of the EMR. These first observations are quite surprising since both models are trained on English data. While we could expect higher performance with French models (CamemBERT for example), the fact that these models are trained on specialized data for one (BioBERT) and on a model designed for the targeted task (SOTA on question-answering for BART) finally shows that language models trained on generic data are inefficient for the MCQA task on medical domain.

In all considered architectures, context seems to have a small impact on systems performance, with a limited increase or drop depending on the configurations. Clearly, the \textbf{RoBERTa} performance is much higher without context ({\it i.e.} without the use of the retriever part), while models based on \textbf{BERT} generally (8 times on 12) outperform their own baseline performances with external context. The fact that we consider the 1-best passage only may explain this impact. 


Concerning \textbf{XLM-RoBERTa-base} (cross lingual representation), we obtain in the case of the context extracted using BM25 from Wikipedia, the worst Hamming score and EMR out of all the discriminative approaches. This confirms our first observation that a non-specialized model does not allow to achieve the best performance on this task.

Using BM25 promotes better context than semantic search using \textbf{MiniLMv2} on both Wikipedia and HAL for most of the runs. Finally, the source depends of the retriever and model used. A majority of the experiments demonstrate that HAL outperforms Wikipedia on BM25 despite the fact that the best model was obtained using Wikipedia.

The scripts to replicate the experiments\footnote{\href{https://github.com/qanastek/FrenchMedMCQA}{https://github.com/qanastek/FrenchMedMCQA}} as well as the pre-trained models\footnote{\href{https://huggingface.co/qanastek/FrenchMedMCQA-BioBERT-V1.1-Wikipedia-BM25/tree/main}{https://huggingface.co/qanastek/FrenchMedMCQA-BioBERT-V1.1-Wikipedia-BM25/tree/main}} are available online.






\section{Conclusion}
\label{sec:conclusion}


We proposed in this paper FrenchMedMCQA, an original, open and publicly available Multiple-Choice Question Answering (MCQA) dataset in the medical field. This is the first French corpus in this domain, including single and multiple answers to questions. 
Several state-the art systems have been evaluated to show current performance on the  dataset. The analysis of these first results notably highlighted the fact that language models specialized to the medical domain allow us to reach better performance than generic models, even if these have been trained in a different language (here, English biomedical models applied to French).

In future works, we will focus on improving the existing methods for the task of MCQA, considering other strategies for the retriever module (multiple passages, combining contexts...). Likewise, we will also consider the construction of data representation models for French specialized for medical domain.



 \section{Acknowledgments}
 \label{sec:acknowledgements}

This work was financially supported by Zenidoc, the DIETS project financed by the Agence Nationale de la Recherche (ANR) under contract ANR-20-CE23-0005 and the ANR AIBy4 (ANR-20-THIA-0011). This work was performed using HPC resources from GENCI-IDRIS (Grant 2022-AD011013061R1 and 2022-AD011013715).

\bibliography{anthology,custom}

\begin{thebibliography}{25}
\expandafter\ifx\csname natexlab\endcsname\relax\def\natexlab#1{#1}\fi

\bibitem[{Clark et~al.(2018)Clark, Cowhey, Etzioni, Khot, Sabharwal, Schoenick,
  and Tafjord}]{Clark2018ThinkYH}
Peter Clark, Isaac Cowhey, Oren Etzioni, Tushar Khot, Ashish Sabharwal, Carissa
  Schoenick, and Oyvind Tafjord. 2018.
\newblock Think you have solved question answering? try arc, the ai2 reasoning
  challenge.
\newblock \emph{ArXiv}, abs/1803.05457.

\bibitem[{Conneau et~al.(2020)Conneau, Khandelwal, Goyal, Chaudhary, Wenzek,
  Guzm{\'a}n, Grave, Ott, Zettlemoyer, and
  Stoyanov}]{conneau-etal-2020-unsupervised}
Alexis Conneau, Kartikay Khandelwal, Naman Goyal, Vishrav Chaudhary, Guillaume
  Wenzek, Francisco Guzm{\'a}n, Edouard Grave, Myle Ott, Luke Zettlemoyer, and
  Veselin Stoyanov. 2020.
\newblock \href {https://doi.org/10.18653/v1/2020.acl-main.747} {Unsupervised
  cross-lingual representation learning at scale}.
\newblock In \emph{Proceedings of the 58th Annual Meeting of the Association
  for Computational Linguistics}, pages 8440--8451, Online. Association for
  Computational Linguistics.

\bibitem[{Grabar et~al.(2018)Grabar, Claveau, and Dalloux}]{grabar2018}
Natalia Grabar, Vincent Claveau, and Cl{\'e}ment Dalloux. 2018.
\newblock \href {https://doi.org/10.18653/v1/W18-5614} {{CAS}: {F}rench corpus
  with clinical cases}.
\newblock In \emph{Proceedings of the Ninth International Workshop on Health
  Text Mining and Information Analysis}, pages 122--128, Brussels, Belgium.
  Association for Computational Linguistics.

\bibitem[{Gu et~al.(2022)Gu, Tinn, Cheng, Lucas, Usuyama, Liu, Naumann, Gao,
  and Poon}]{Gu_2022}
Yu~Gu, Robert Tinn, Hao Cheng, Michael Lucas, Naoto Usuyama, Xiaodong Liu,
  Tristan Naumann, Jianfeng Gao, and Hoifung Poon. 2022.
\newblock \href {https://doi.org/10.1145/3458754} {Domain-specific language
  model pretraining for biomedical natural language processing}.
\newblock \emph{{ACM} Transactions on Computing for Healthcare}, 3(1):1--23.

\bibitem[{Izacard and Grave(2020)}]{izacard2020leveraging}
Gautier Izacard and Edouard Grave. 2020.
\newblock Leveraging passage retrieval with generative models for open domain
  question answering.
\newblock \emph{arXiv preprint arXiv:2007.01282}.

\bibitem[{Khashabi et~al.(2020)Khashabi, Min, Khot, Sabharwal, Tafjord, Clark,
  and Hajishirzi}]{https://doi.org/10.48550/arxiv.2005.00700}
Daniel Khashabi, Sewon Min, Tushar Khot, Ashish Sabharwal, Oyvind Tafjord,
  Peter Clark, and Hannaneh Hajishirzi. 2020.
\newblock \href {https://doi.org/10.48550/ARXIV.2005.00700} {Unifiedqa:
  Crossing format boundaries with a single qa system}.

\bibitem[{Khot et~al.(2019)Khot, Clark, Guerquin, Jansen, and
  Sabharwal}]{https://doi.org/10.48550/arxiv.1910.11473}
Tushar Khot, Peter Clark, Michal Guerquin, Peter Jansen, and Ashish Sabharwal.
  2019.
\newblock \href {https://doi.org/10.48550/ARXIV.1910.11473} {Qasc: A dataset
  for question answering via sentence composition}.

\bibitem[{Lai et~al.(2017)Lai, Xie, Liu, Yang, and Hovy}]{lai-etal-2017-race}
Guokun Lai, Qizhe Xie, Hanxiao Liu, Yiming Yang, and Eduard Hovy. 2017.
\newblock \href {https://doi.org/10.18653/v1/D17-1082} {{RACE}: Large-scale
  {R}e{A}ding comprehension dataset from examinations}.
\newblock In \emph{Proceedings of the 2017 Conference on Empirical Methods in
  Natural Language Processing}, pages 785--794, Copenhagen, Denmark.
  Association for Computational Linguistics.

\bibitem[{Lee et~al.(2019)Lee, Yoon, Kim, Kim, Kim, So, and Kang}]{Lee_2019}
Jinhyuk Lee, Wonjin Yoon, Sungdong Kim, Donghyeon Kim, Sunkyu Kim, Chan~Ho So,
  and Jaewoo Kang. 2019.
\newblock \href {https://doi.org/10.1093/bioinformatics/btz682} {{BioBERT}: a
  pre-trained biomedical language representation model for biomedical text
  mining}.
\newblock \emph{Bioinformatics}.

\bibitem[{Lewis et~al.(2019)Lewis, Liu, Goyal, Ghazvininejad, Mohamed, Levy,
  Stoyanov, and Zettlemoyer}]{https://doi.org/10.48550/arxiv.1910.13461}
Mike Lewis, Yinhan Liu, Naman Goyal, Marjan Ghazvininejad, Abdelrahman Mohamed,
  Omer Levy, Ves Stoyanov, and Luke Zettlemoyer. 2019.
\newblock \href {https://doi.org/10.48550/ARXIV.1910.13461} {Bart: Denoising
  sequence-to-sequence pre-training for natural language generation,
  translation, and comprehension}.

\bibitem[{Li et~al.(2021)Li, Zhong, and Chen}]{li-etal-2021-mlec}
Jing Li, Shangping Zhong, and Kaizhi Chen. 2021.
\newblock \href {https://aclanthology.org/2021.emnlp-main.698} {{MLEC-QA}: {A}
  {C}hinese {M}ulti-{C}hoice {B}iomedical {Q}uestion {A}nswering {D}ataset}.
\newblock In \emph{Proceedings of the 2021 Conference on Empirical Methods in
  Natural Language Processing}, pages 8862--8874, Online and Punta Cana,
  Dominican Republic. Association for Computational Linguistics.

\bibitem[{Liu et~al.(2019)Liu, Ott, Goyal, Du, Joshi, Chen, Levy, Lewis,
  Zettlemoyer, and Stoyanov}]{https://doi.org/10.48550/arxiv.1907.11692}
Yinhan Liu, Myle Ott, Naman Goyal, Jingfei Du, Mandar Joshi, Danqi Chen, Omer
  Levy, Mike Lewis, Luke Zettlemoyer, and Veselin Stoyanov. 2019.
\newblock \href {https://doi.org/10.48550/ARXIV.1907.11692} {Roberta: A
  robustly optimized bert pretraining approach}.

\bibitem[{Martin et~al.(2020)Martin, Muller, Ortiz~Su{\'a}rez, Dupont, Romary,
  de~la Clergerie, Seddah, and Sagot}]{martin-etal-2020-camembert}
Louis Martin, Benjamin Muller, Pedro~Javier Ortiz~Su{\'a}rez, Yoann Dupont,
  Laurent Romary, {\'E}ric de~la Clergerie, Djam{\'e} Seddah, and Beno{\^\i}t
  Sagot. 2020.
\newblock \href {https://www.aclweb.org/anthology/2020.acl-main.645}
  {{C}amem{BERT}: a tasty {F}rench language model}.
\newblock In \emph{Proceedings of the 58th Annual Meeting of the Association
  for Computational Linguistics}, pages 7203--7219, Online. Association for
  Computational Linguistics.

\bibitem[{Mihaylov et~al.(2018)Mihaylov, Clark, Khot, and
  Sabharwal}]{OpenBookQA2018}
Todor Mihaylov, Peter Clark, Tushar Khot, and Ashish Sabharwal. 2018.
\newblock Can a suit of armor conduct electricity? a new dataset for open book
  question answering.
\newblock In \emph{EMNLP}.

\bibitem[{Névéol et~al.(2014)Névéol, Grouin, Leixa, Rosset, and
  Zweigenbaum}]{neveol14quaero}
Aurélie Névéol, Cyril Grouin, Jeremy Leixa, Sophie Rosset, and Pierre
  Zweigenbaum. 2014.
\newblock The {QUAERO} {French} medical corpus: A ressource for medical entity
  recognition and normalization.
\newblock In \emph{Proc of BioTextMining Work}, pages 24--30.

\bibitem[{Pal et~al.(2022)Pal, Umapathi, and Sankarasubbu}]{pmlr-v174-pal22a}
Ankit Pal, Logesh~Kumar Umapathi, and Malaikannan Sankarasubbu. 2022.
\newblock \href {https://proceedings.mlr.press/v174/pal22a.html} {Medmcqa: A
  large-scale multi-subject multi-choice dataset for medical domain question
  answering}.
\newblock In \emph{Proceedings of the Conference on Health, Inference, and
  Learning}, volume 174 of \emph{Proceedings of Machine Learning Research},
  pages 248--260. PMLR.

\bibitem[{Radford et~al.(2019)Radford, Wu, Child, Luan, Amodei, and
  Sutskever}]{Radford2019LanguageMA}
Alec Radford, Jeff Wu, Rewon Child, David Luan, Dario Amodei, and Ilya
  Sutskever. 2019.
\newblock Language models are unsupervised multitask learners.

\bibitem[{Reimers and
  Gurevych(2019)}]{https://doi.org/10.48550/arxiv.1908.10084}
Nils Reimers and Iryna Gurevych. 2019.
\newblock \href {https://doi.org/10.48550/ARXIV.1908.10084} {Sentence-bert:
  Sentence embeddings using siamese bert-networks}.

\bibitem[{Richardson et~al.(2013)Richardson, Burges, and
  Renshaw}]{richardson-etal-2013-mctest}
Matthew Richardson, Christopher~J.C. Burges, and Erin Renshaw. 2013.
\newblock \href {https://aclanthology.org/D13-1020} {{MCT}est: A challenge
  dataset for the open-domain machine comprehension of text}.
\newblock In \emph{Proceedings of the 2013 Conference on Empirical Methods in
  Natural Language Processing}, pages 193--203, Seattle, Washington, USA.
  Association for Computational Linguistics.

\bibitem[{Robertson and Sparck~Jones(1988)}]{10.5555/106765.106783}
Stephen~E. Robertson and Karen Sparck~Jones. 1988.
\newblock \emph{Relevance Weighting of Search Terms}, page 143–160. Taylor
  Graham Publishing, GBR.

\bibitem[{Sap et~al.(2019)Sap, Rashkin, Chen, LeBras, and
  Choi}]{https://doi.org/10.48550/arxiv.1904.09728}
Maarten Sap, Hannah Rashkin, Derek Chen, Ronan LeBras, and Yejin Choi. 2019.
\newblock \href {https://doi.org/10.48550/ARXIV.1904.09728} {Socialiqa:
  Commonsense reasoning about social interactions}.

\bibitem[{Trotman et~al.(2014)Trotman, Puurula, and
  Burgess}]{10.1145/2682862.2682863}
Andrew Trotman, Antti Puurula, and Blake Burgess. 2014.
\newblock \href {https://doi.org/10.1145/2682862.2682863} {Improvements to bm25
  and language models examined}.
\newblock In \emph{Proceedings of the 2014 Australasian Document Computing
  Symposium}, ADCS '14, page 58–65, New York, NY, USA. Association for
  Computing Machinery.

\bibitem[{Vilares and
  G{\'o}mez-Rodr{\'\i}guez(2019)}]{vilares-gomez-rodriguez-2019-head}
David Vilares and Carlos G{\'o}mez-Rodr{\'\i}guez. 2019.
\newblock \href {https://doi.org/10.18653/v1/P19-1092} {{HEAD}-{QA}: A
  healthcare dataset for complex reasoning}.
\newblock In \emph{Proceedings of the 57th Annual Meeting of the Association
  for Computational Linguistics}, pages 960--966, Florence, Italy. Association
  for Computational Linguistics.

\bibitem[{Wang et~al.(2020)Wang, Bao, Huang, Dong, and
  Wei}]{https://doi.org/10.48550/arxiv.2012.15828}
Wenhui Wang, Hangbo Bao, Shaohan Huang, Li~Dong, and Furu Wei. 2020.
\newblock \href {https://doi.org/10.48550/ARXIV.2012.15828} {Minilmv2:
  Multi-head self-attention relation distillation for compressing pretrained
  transformers}.

\bibitem[{Wolf et~al.(2019)Wolf, Debut, Sanh, Chaumond, Delangue, Moi, Cistac,
  Rault, Louf, Funtowicz, Davison, Shleifer, von Platen, Ma, Jernite, Plu, Xu,
  Scao, Gugger, Drame, Lhoest, and
  Rush}]{https://doi.org/10.48550/arxiv.1910.03771}
Thomas Wolf, Lysandre Debut, Victor Sanh, Julien Chaumond, Clement Delangue,
  Anthony Moi, Pierric Cistac, Tim Rault, Rémi Louf, Morgan Funtowicz, Joe
  Davison, Sam Shleifer, Patrick von Platen, Clara Ma, Yacine Jernite, Julien
  Plu, Canwen Xu, Teven~Le Scao, Sylvain Gugger, Mariama Drame, Quentin Lhoest,
  and Alexander~M. Rush. 2019.
\newblock \href {https://doi.org/10.48550/ARXIV.1910.03771} {Huggingface's
  transformers: State-of-the-art natural language processing}.

\end{thebibliography}




\end{document}